\documentclass[10pt, reqno, a4paper]{amsart}

\usepackage[T1]{fontenc}
\usepackage{lmodern}
\usepackage[english]{babel}
\usepackage{amsfonts}
\usepackage{amsmath} 
\usepackage{amssymb}
\usepackage{amsthm}
\usepackage{mathrsfs}
\usepackage[x11names,dvipsnames,svgnames]{xcolor}
\usepackage{booktabs}
\usepackage{float}
\usepackage{enumitem}
\usepackage{graphicx}
\usepackage[all]{xy}
\usepackage{multirow}
\usepackage{array}
\usepackage{adjustbox}
\usepackage{calc}
\usepackage{algorithm}
\usepackage{algpseudocode} 
\RequirePackage{doi}
\usepackage{hyperref}
\hypersetup{
    colorlinks,
    linkcolor={red!50!black},
    citecolor={blue!50!black},
    urlcolor={blue!80!black}
}

\usepackage[noabbrev]{cleveref}
\usepackage{bm}

\algrenewcommand\algorithmicwhile{\textbf{While}}
\algrenewcommand\algorithmicfor{\textbf{For}}
\algrenewcommand\algorithmicdo{\textbf{Do}}
\algrenewcommand\algorithmicif{\textbf{If}}
\algrenewcommand\algorithmicthen{\textbf{Then}}
\algrenewcommand\algorithmicelse{\textbf{Else}}
\algrenewcommand\algorithmicend{\textbf{End}}
\algrenewcommand\algorithmicreturn{\textbf{Return}}

\theoremstyle{plain}
\newtheorem{lemma}{Lemma}[section]

\theoremstyle{definition}

\newtheorem{remark}[lemma]{Remark}

\newcommand{\R}{\mathbb{R}}
\newcommand{\C}{\mathbb{C}}
\newcommand{\p}{\mathbb{P}}

\newcommand{\id}{\mathrm{id}}
\newcommand{\SE}{\mathrm{SE}_3}

\newcommand{\inv}{\mathrm{inv}}
\newcommand{\pinv}{\mathrm{pinv}}

\newcommand{\BSC}{\mathrm{BSC}}

\newcommand{\mscr}{\mathscr}

\title{A new line-symmetric mobile infinity-pod}
\date{}

\begin{document}

\author[Gallet]{Matteo Gallet$^\ast$}
\address[MG]{Radon Institute for Computational and Applied Mathematics (RICAM),
Altenbergerstra\ss e 69 \\
4040 Linz, Austria}
\thanks{$^\ast$ Supported by the Austrian Science Fund (FWF): Erwin Schr\"odinger Fellowship J4253.}
\email{matteo.gallet@ricam.oeaw.ac.at}

\author[Schicho]{Josef Schicho$^\circ$}
\address[JS]{Research Institute for Symbolic Computation (RISC), Johannes Kepler University Linz,
Altenbergerstra\ss e 69 \\ 
4040 Linz, Austria}
\thanks{$^\circ$ Supported by the Austrian Science Fund (FWF): W1214-N15, project DK9.}
\email{jschicho@risc.jku.at}

\begin{abstract}
 We construct parallel manipulators with one degree of freedom and admitting infinitely many legs lying on a curve of degree ten and genus six.
 Our technique relies upon a duality between the spaces parametrizing all the possible legs and all the possible configurations of a manipulator.
 Before describing our construction, we show how this duality helps explaining several known phenomena regarding mobility of parallel manipulators.
\end{abstract}

\maketitle

\section{Introduction}

Parallel manipulators, or pods, are mechanical devices constituted of two rigid bodies, called the \emph{base} and the \emph{platform},
that are connected by rigid rods, called \emph{legs}. Each leg is anchored to the base and the platform via \emph{spherical joints},
namely joints that allow arbitrary rotations around the anchor point.
Often, there is the possibility to change the length of the legs in order to take the platform to a given pose with respect to the base.
It is then undesirable that the platform may change its pose with respect to the base
without changing the leg lengths.
This phenomenon is called a \emph{self-motion} of the manipulator.
The study of self-motions of parallel manipulators has a long history
and has benefited since the beginning from algebraic and geometric techniques. 
In this paper, we focus on \emph{mobile infinity-pods}, namely parallel manipulators that admit self-motions with an infinite number of legs.
One can easily produce examples of mobile infinity-pods by taking congruent base and platform anchor points, with legs of the same length,
or by having all base or platform points aligned.
There are also other examples of mobile infinity-pods, 
as the one discovered by Husty and Karger in \cite{Karger1998, Karger2008a},
or those described in \cite{Nawratil2011, Nawratil2013, Husty2002b}.

We outline a construction of a family of mobile infinity-pods which is, to our knowledge, not present in the literature.
These mobile infinity-pods are \emph{line-symmetric}; this means the following. 
A possible pose, or \emph{configuration}, of the platform of a parallel manipulator may be encoded via the direct isometry that maps it to a given initial pose.
Therefore, the set of possible configurations of a parallel manipulator can be encoded as a subset of the group of direct isometries of the space.
A line-symmetric mobile pod is a parallel manipulator for which all its configurations are given by involutions,
namely, half-turns around lines, possibly composed with a fixed isometry.
For more information about line-symmetric motions, one can see~\cite{Krames1937, Gallet2017, Nawratil2018}.

Our construction relies on a particular choice of a projective model of the group of direct isometries.
As a consequence of this choice, the condition imposed on direct isometries, i.e., on configurations of the pod, by the presence of a leg is linear.
Moreover, one can define a projective model of the space parametrizing all possible legs
and see that there is a duality between the ambient space of the projective model of the isometries
and the ambient space of the projective model of the legs.
The interplay between isometries and legs allowed by this duality, established in~\cite{Gallet2017},
is the backbone of our work.

Before describing our construction, we get acquainted with this setting.
We examine, in the light of the duality between legs and isometries,
the well known result by Duporcq that a planar pentapod can be extended to a hexapod
without changing its set of possible configurations.
After that, we investigate a construction of planar line-symmetric mobile infinity-pods
whose base and platform anchor points belong to a cubic curve.
These infinity-pods have been studied by Husty and Karger via a more algebraic method \cite{Karger1998}.
We mainly argue via geometric reasoning, rather than relying on algebraic computations.
In discussing this case, we re-prove classical results obtained by Borel and Bricard.
We believe that these examples indicate the versatility of this way of studying parallel manipulators.

Once we familiarized with going back and forth between legs and isometries,
we explain our main construction, which starts from a curve in three-dimensional projective space,
representing the rotational parts of the configurations of our desired infinity-pod.
The curve is constructed in such a way that it can be lifted to the projective model of direct isometries
(more precisely, to the subvariety parametrizing involutions)
and it has a linear span of the ``right'' dimension.
This ensures that the space of legs compatible with such a curve obtained by duality is one-dimensional,
namely we obtain a line-symmetric mobile infinity-pod.
The base and platform points of such a line-symmetric mobile infinity-pod
lie on a curve of degree ten and genus six.
These examples are, to our knowledge, new.

\Cref{projective_models} and \Cref{leg_spaces} recall the projective models of isometries and leg spaces that we use in our construction.
\Cref{duporcq} serves as a warm-up to get acquainted with the setup introduced in the previous sections
and re-proves the well known result stating that we can add a leg to a planar pentapod without changing its configuration space.
\Cref{cubic} describes line-symmetric mobile infinity-pods whose base and platform points are planar and lie on a cubic curve.
\Cref{construction} explains our main result, namely, how to construct line-symmetric mobile infinity-pods
whose base and platform points lie on a space curve of degree ten.
We provide some examples in the ancillary Macaulay2~\cite{M2} files \texttt{InfinityPods.m2} and \texttt{CubicExample.m2} available at the arXiv page of this paper.

{\bfseries Acknowledgments.} We thank Georg Nawratil for providing us with useful information about the known examples of infinity-pods.

\section{Compactifications of space isometries}
\label{projective_models}

We describe a projective model of the group~$\SE$ of isometries of~$\R^3$ and another one for the subgroup of involutions.
No new results are present here, this section is only meant for self-containedness.

To define the projective model of isometries we use, 
introduced in~\cite{Gallet2015} and in~\cite[Section~5]{Mourrain1996},
we consider the map that sends an isometry~$\sigma = (M, y)$ with rotational part~$M$ and translational part~$y$ to the point in~$\p^{16}$
\[
 (M: \underbrace{-M^{t}y}_{:=x}: y: \underbrace{\left\langle y, y \right\rangle}_{:=r}: \underbrace{1}_{:=h}) \,.
\]
This means that the points in~$\p^{16}$ corresponding to elements in~$\SE$ satisfy
\begin{equation*}
\begin{gathered}
 M M^{t} = M^{t} M = h^2 \cdot \id_{\R^3} \,, \quad \det(M) = h^3 \,, \\
 Mx + hy = 0 \,, \quad M^{t}y + hx = 0 \,, \\
 rh = \left\langle x, x \right\rangle = \left\langle y, y \right\rangle \,, \quad h \neq 0 \,.
\end{gathered}
\end{equation*}
We denote by~$X$ the Zariski closure of the image of~$\SE$ in~$\p^{16}$.
In this way, the variety~$X$, which is of dimension six and degree forty,
contains a copy of~$\SE$ as an open subset.
The complement of this~$\SE$ is the locus of points in~$X$ for which $h = 0$,
which is a cone over the point~$(0: \dotsb : 1 : 0)$.

The condition imposed on isometries $\sigma \in \SE$ by the existence of a leg with base point~$a$,
platform point~$b$, and length~$d$, namely
\[
 \left\| \sigma(a) - b \right\|^2 = d^2
\]
reads, in these coordinates,
\begin{equation}
\label{eq:sphere_condition}
 \bigl( 
  \left\langle a, a \right\rangle + \left\langle b, b \right\rangle - d^2
 \bigr) h + r -
 2 \left\langle a, x \right\rangle - 
 2 \left\langle b, x \right\rangle - 
 2 \left\langle M a, b \right\rangle = 0 \,.
\end{equation}
We call this equation the \emph{sphere condition} imposed by the leg $(a,b,d)$.
Notice that the latter is a \emph{linear} equation in the coordinates of~$\p^{16}$.
Therefore, the (projective model of the) space of configurations of a pod,
considered as a subset of the group of isometries, 
is always of the form $X \cap \Lambda$, where $\Lambda$ is a linear space.

Projecting~$X$ to the $(M: h)$-coordinates yields a projective model of the group~$\mathrm{SO}_3$ of rotations.
We denote by~$X_{\mathrm{rot}} \subset \p^9$ the (Zariski closure of the) image of such a projection.
We know that, up to an automorphism of~$\p^9$, the variety~$X_{\mathrm{rot}}$ is
the image of~$\p^3$ under the Veronese map defined by quadratic monomials:
\begin{equation}
\label{eq:veronese}
 \begin{array}{rcl}
  \p^3 & \longrightarrow & \p^9 \\ 
  (e_0: e_1: e_2: e_3) & \mapsto & 
  \begin{array}{c} (e_0^2: e_1^2: e_2^2: e_3^2: \\[-2pt] e_0 e_1: e_0 e_2: e_0 e_3: e_1 e_2: e_1 e_3: e_2 e_3) \end{array}
 \end{array}
\end{equation}

We now focus on involutions in~$\SE$, 
namely rotations by~$180^{\circ}$ around an axis.
As described in~\cite{Gallet2017}, if we impose on~$X$ the equations
\[
 M = M^{t} \quad \text{and} \quad x = y \,,
\]
then we obtain a set with two irreducible components:
one isolated point corresponding to the identity of~$\SE$
and a four-dimensional variety satisfying the further equation
\[
 m_{11} + m_{22} + m_{33} + h = 0 \,.
\]
We call the latter variety~$X_{\inv}$, which is then a projective model of involutions in~$\SE$.

As described in \cite[Section~3]{Gallet2017}, 
we can provide another projective model for involutions in~$\SE$ as follows.
We define $X_1$ to be the subvariety
\[
 X_1 :=
 \bigl\{
  (M:x:y:r:h) \in X
  \; \mid \;
  m_{ij} = m_{ji}
  \  \text{and} \ 
  m_{11} + m_{22} + m_{33} + h = 0
 \bigr\} \,.
\]
Notice that $X_{\inv} = X_1 \cap \{ x = y \}$.
We project $X_1$ from the point $\{ M = x = y = h = 0 \}$ obtaining $X_2 \subset \p^{15}$.
If we express the $(M,h)$-coordinates of the points of~$X_2$ in terms of the $(e_0: e_1: e_2: e_3)$-coordinates 
by inverting the Veronese map, we see that it holds $e_0 = 0$. The three remaining $e$-coordinates,
together with 
\[
\begin{gathered}
 p_i := x_i + y_i \quad \text{for } i \in \{1,2,3\} \,, \\
 q_i := x_i - y_i \quad \text{for } i \in \{1,2,3\} \,, \\
\end{gathered}
\]
define a map
\[
 X_2 \longrightarrow \p(\vec{1}, \vec{2}) \,,
\]
where $\vec{1} = (1,1,1)$, $\vec{2} = (2,2,2,2,2,2)$, 
and $\p(\vec{1}, \vec{2})$ is the weighted projective space in which the $e$-coordinates have degree one, 
while the $p$- and $q$-coordinates have degree two. 
The image of~$X_2$ under this map is a variety $Z \subset \p(\vec{1}, \vec{2})$ of dimension five, 
defined by the equations
\[
\begin{gathered}
 e_1 p_1 + e_2 p_2 + e_3 p_3 = 0 \,, \\
 p_1 q_1 + p_2 q_2 + p_3 q_3 = 0 \,, \\
 e_1 q_2 - e_2 q_1 = e_1 q_3 - e_3 q_1 = e_2 q_3 - e_3 q_2 = 0 \,.
\end{gathered}
\]
Since $X_{\inv}$ is contained in~$X_1$, we can consider its image in~$Z$, which we call~$Z_{\inv}$ and is given by
\begin{equation}
\label{eq:Zinv}
\begin{aligned}
 Z_{\inv} &= Z \cap \{ q_1 = q_2 = q_3 = 0 \} \\
          &= \bigl\{ (e: p: q) \in \p(\vec{1}, \vec{2}) \; \mid \; e_1 p_1 + e_2 p_2 + e_3 p_3 = 0 \ \text{and}\ q = 0 \bigr\} \,.
\end{aligned}
\end{equation}

\section{Leg spaces and dualities}
\label{leg_spaces}

In this section, we describe a variety~$Y \subset \p^{16}$ that serves as projective model for the set of all possible legs of parallel manipulators.
In doing that, we establish a duality between the $\p^{16}$ containing the projective model~$X$ of the isometries 
and the $\p^{16}$ containing the projective model~$Y$ of all possible legs. 
We then explain how this duality restricts to the linear space containing the variety~$X_{\inv}$ of involutions.
As in \Cref{projective_models}, no new results are present here; this section is only meant for self-containedness and reports material from \cite{Gallet2017}.

We start from \Cref{eq:sphere_condition} and we define the \emph{corrected leg length} to be the quantity
\[
 l := \left\langle a, a \right\rangle + \left\langle b, b \right\rangle - d^2 \,.
\]
After that, we think of the base point $a= (a_1, a_2, a_3)$ and the platform point $b = (b_1, b_2, b_3)$
as points in~$\p^3$ with coordinates $(1: a_1: a_2: a_3)$ and $(1: b_1: b_2: b_3)$.
We then consider the Segre embedding $\p^3 \times \p^3 \hookrightarrow \p^{15}$ given by
\[
 (a_0: a_1: a_2: a_3) \times (b_0: b_1: b_2: b_3) \mapsto (a_0 b_0: \dotsb: \underbrace{a_i b_j}_{:=z_{ij}}: \dotsb: a_3 b_3) \,.
\]
In the new variables $(\{z_{ij}\}: l)$, the sphere condition from \Cref{eq:sphere_condition} becomes
\begin{equation}
\label{eq:bilinear_sphere_condition}
  l  h + z_{00} r  - 2  ( z_{10} x_1 + z_{20} x_2 + z_{30} x_3 ) 
  - 2  (z_{01} y_1 + z_{02} y_2 + z_{03} y_3 ) - 2 
  \sum_{i, j = 1}^{3} m_{ij}  z_{ij} =  0 \,.
\end{equation}
What we got is a \emph{bilinear} form in the coordinates $(M: x: y: r: h)$ and $(\{z_{ij}\}: l)$
between the~$\p^{16}$ containing the variety~$X$ and the~$\p^{16}$ containing the cone with vertex $\{ z_{ij} = 0 \text{ for all } i,j \,,\, l = 1\}$ over the Segre variety in~$\p^{15}$.
We call this form the \emph{bilinear sphere condition} and we denote it by $\BSC(M,hx,y,r,h,z,l)$.
We denote the cone over the Segre variety by~$Y$; it is a projective model of the set of all possible legs.
The duality provided by the bilinear sphere condition allows us to go back and forth between configurations and legs as follows:
\begin{itemize}
 \item Given a point $\overline{\sigma} = (\overline{M}: \overline{x}: \overline{y}: \overline{r}: \overline{h})$ in~$\SE \subset X$, 
 let $L_{\overline{\sigma}}$ be the hyperplane of equation $\BSC(\overline{\sigma}, z, l) = 0$.
 Then the intersection $Y \cap L_{\overline{\sigma}}$ is the (projective model of the) set of legs $(a,b,d)$ such that $\left\| \overline{\sigma}(a) - b \right\|^2 = d^2$.
 \item Given a leg $\overline{\lambda} = (\overline{a}, \overline{b}, \overline{d})$ in $\R^3 \times \R^3 \times \R \subset Y$ 
 with coordinates $(\overline{z}: \overline{l})$, let $K_{\overline{\lambda}}$ be the hyperplane of equation $\BSC(M, x, y, r, h, \overline{z}, \overline{l}) = 0$.
 Then the intersection $X \cap K_{\overline{\lambda}}$ is the (projective model of the) set of isometries~$\sigma$ 
 such that $\left\| \sigma(\overline{a}) - \overline{b} \right\| = \overline{d}^2$.
\end{itemize}

We can restrict the bilinear sphere condition to the~$\p^{10}$ defined by the equations $M = M^{t}$ and $x = y$, namely the~$\p^{10}$ where $X_{\inv}$ lives.
By using these equations and setting
\[
 s_{ij} := z_{ij} + z_{ji} \quad \text{for any } i < j \,,
\]
one gets that the bilinear sphere condition becomes
\begin{equation}
\label{eq:symmetric_bilinear_sphere_condition}
 lh + z_{00} r - 2 \sum_{i=1}^{3} s_{0i} x_i - 2 \sum_{i=1}^{3} z_{ii} m_{ii} - 2 \!\! \sum_{1 \leq i < j \leq 3} s_{ij} m_{ij} = 0 \,.
\end{equation}
We call this form the \emph{symmetric bilinear sphere condition}.
The symmetric bilinear sphere condition determines a duality between the $\p^{10}$ with coordinates $(\{m_{ij}\}_{i \leq j} :x:r:h)$ 
and the $\p^{10}$ with coordinates $(z_{11}: z_{22}: z_{33}: s_{12}: s_{13}: s_{23}: s_{01}: s_{02}: s_{03}: z_{00}: l)$, 
which is a projection of the $\p^{16}$ containing~$Y$.
The projection of~$Y$ onto the latter coordinates is the cone over the image of the map
\begin{equation}
\label{eq:quotient_legs}
 \begin{array}{rrcl}
  \alpha \colon & \p^3 \times \p^3 & \longrightarrow & \p^{9} \\
  & \begin{array}{c} (a_0: a_1: a_2: a_3) \\[-2pt] \times \\[-2pt] (b_0: b_1: b_2: b_3) \end{array} & \mapsto & 
    (a_1 b_1: a_2 b_2: \dotsb: \underbrace{a_i b_j + a_j b_i}_{s_{ij}}: \dotsb: a_0 b_0)
 \end{array}
\end{equation}
We denote this cone by~$Y_{\inv}$: it parametrizes pairs of legs obtained by swapping base and platform anchor points\footnote{Here, 
differently from \cite{Gallet2017}, we denote by~$Y_{\inv}$ the whole cone and not just the image of the map~$\alpha$.}.
The variety~$Y_{\inv}$ has dimension~seven and degree~ten.

\section{Warming up: Duporcq's result}
\label{duporcq}

To get acquainted to the setup introduced in \Cref{projective_models} and \Cref{leg_spaces},
we discuss the well known result by Duporcq \cite{Duporcq1898} which, roughly speaking, states that we can add
a leg to a mobile planar pentapod without reducing its mobility.
We will be, on purpose, not very precise in handling all the possible ``extreme'' cases of Duporcq's statement.
Our goal here is, in fact, only to showcase how we can use the duality between isometries and legs
to understand the phenomenon.
We refer to \cite{Nawratil2014} for a thorough and precise analysis of Duporcq's result.

To start, we specialize the bilinear sphere equation from \Cref{eq:bilinear_sphere_condition} 
to the setting where base and platform are planar.
Therefore, in~$Y$ we consider the cone~$Y_{\mathrm{p}}$ (where ``$\mathrm{p}$'' stands for ``planar'') over the image of
\[
 \{ a_3 = 0 \} \times \{ b_3 = 0 \} 
 \; \cong \;
 \p^2 \times \p^2
\]
under the Segre embedding. This cone is given by
\[
 Y \cap \{ z_{03} = z_{13} = z_{23} = z_{33} = z_{32} = z_{31} = z_{30} = 0 \}
\]
and is therefore contained in the~$\p^{9}$ with coordinates $(\{z_{ij}\}_{0 \leq i \leq  j \leq 2}: l)$.
The variety~$Y_{\mathrm{p}}$ is of dimension five and degree six, 
since the Segre embedding of~$\p^2 \times \p^2$ in~$\p^8$ has dimension four and degree six.
Once restricted to this~$\p^{9}$, the bilinear sphere condition reads
\begin{equation}
\label{eq:planar_bilinear_sphere_condition}
  l  h + z_{00} r  - 2  ( z_{10} x_1 + z_{20} x_2 ) 
  - 2  (z_{01} y_1 + z_{02} y_2 + z_{03}) - 2 
  \sum_{i, j = 1}^{2} m_{ij}  z_{ij} =  0 \,,
\end{equation}
hence it provides a duality between the~$\p^9$ containing~$Y_{\mathrm{p}}$ and the~$\p^{9}$ with coordinates 
\[
 (\{m_{ij}\}_{1 \leq i,j \leq 2}: x_1: x_2: y_1: y_2: r: h) \,.
\]
The projection of~$X$ onto the latter coordinates is a variety of dimension six and degree twenty, which we denote by~$X_{\mathrm{p}}$.

A planar pentapod is determined by five legs, namely five points in~$Y_{\mathrm{p}}$.
In the general case, these five points span a~$\p^4$.
The dual linear space to these five points is then a~$\p^4$ in the~$\p^9$ containing~$X_{\mathrm{p}}$,
which hence intersects~$X_{\mathrm{p}}$ is a curve.
So we recovered that a general planar pentapod is mobile 
(which is well known and is actually true for any general pentapod, without the planarity assumption).
But we can say something more: since $Y_{\mathrm{p}}$ is five-dimensional of degree six,
the~$\p^4$ generated by the legs of the pentapod will intersect it in a sixth point.
However, the dual space to these six points coincides with the dual space to the original five points,
since by construction it depends only on the span of the points.
Thus, the configurations of the new hexapod coincide with the configurations of the original pentapod.

\begin{remark}
 In Duporcq's construction, the fact that the pentapod is mobile follows from the property that the legs span a~$\p^4$.
 As we pointed out, a \emph{general}~$\p^4$ intersects~$Y_{\mathrm{p}}$ in finitely many points.
 However, a \emph{special} choice of~$\p^4$ may lead to infinitely many legs.
 We owe the following example of this situation to Georg Nawratil.
 Take any two parametrizations of two plane conics $f, g \colon \p^1 \longrightarrow \p^2$ 
 and form the product $h \colon \p^1 \longrightarrow \p^2 \times \p^2$.
 The map~$h$ has hence the form
 \[
  (s: t) \mapsto \bigl( x_0(s,t): x_1(s,t): x_2(s:t) \bigr) \times \bigl( y_0(s,t): y_1(s,t): y_2(s:t) \bigr) \,,
 \]
 where $x_i$ and~$y_j$ are quadratic polynomials for $i, j \in \{0, 1, 2\}$.
 The composition of~$h$ with the Segre embedding gives a map whose components are homogeneous polynomials of degree four in~$s$ and~$t$.
 Let $L$ be the image of~$h$ under the Segre embedding, which is a curve in~$\p^8$.
 Since the linear space of homogeneous bivariate polynomials of degree four is five-dimensional,
 the span of~$L$ is at most a~$\p^4$ and it is exactly a~$\p^4$
 when $f$ and $g$ are general parametrizations.
 Now, it is enough to take any lift~$\widetilde{L}$ of~$L$ to~$Y_{\mathrm{p}}$, 
 which is a cone over the Segre embedding of~$\p^2 \times \p^2$, whose span is also a~$\p^4$.
 This is obtained by simply imposing a general linear equation involving the variable~$l$.
 As in Duporcq's example, the dual of~$\widetilde{L}$ is a~$\p^4$,
 which hence intersects~$X_{\mathrm{p}}$ in a curve, yielding a planar mobile infinity-pod.
\end{remark}

\begin{remark}
 With similar arguments as the ones in this section,
 we can analyze the result of~\cite{Husty2002b},
 which says that, except for some special situations,
 we can add infinitely many legs to a planar hexapod without changing its possible configurations.
 In fact, a planar hexapod determines six points on~$Y_{\mathrm{p}}$,
 which in general span a~$\p^5$.
 Since $Y_{\mathrm{p}}$ has dimension five in~$\p^9$,
 intersecting it with a~$\p^5$ yields, in addition to the six initial points, a whole curve.
 Since the span of this bigger set is the same as the one of the initial six points,
 the dual linear space is the same, hence the possible configurations do not change
 if we add the infinitely many legs belonging to the curve in~$Y_{\mathrm{p}}$.
\end{remark}

\section{Cubic line-symmetric mobile infinity-pods}
\label{cubic}

While studying the possible self-motions of a Stewart-Gough platform,
Husty and Karger discovered a mobile infinity-pod whose base and platform points
are planar and lie on a cubic curve 
and such that the rotational part of its configuration curve is constituted of involutions; 
see \cite{Karger1998, Karger2008a, Karger2008b}.
They derived the existence of this kind of infinity-pods by using the so-called \emph{Study coordinates}
to encode isometries; see \cite{Selig1996}.
Their technique involves a computer algebra manipulation of the conditions imposed on isometry by the legs.
We propose another derivation of line-symmetric mobile infinity-pods via a geometric argument 
employing the projective models and dualities we introduced so far. 
Nawratil in~\cite{Nawratil2011, Nawratil2013} proved that the construction by Husty and Karger is line-symmetric,
so it coincides with ours.
Moreover, our considerations match with those by Borel and Bricard in their famous papers submitted for the Prix Vaillant \cite{Borel1908, Bricard1906}.

First of all, as we did in \Cref{duporcq},
we adapt the symmetric bilinear sphere condition introduced in \Cref{leg_spaces} 
to the situation of planar base and platform points. 
Therefore, we look at the image of 
\[
 \{ a_3 = 0 \} \times \{ b_3 = 0 \} 
 \; \cong \;
 \p^2 \times \p^2
\]
under the map~$\alpha$ from \Cref{eq:quotient_legs}.
What we get is
\[
\begin{gathered}
 (a_0: a_1: a_2: 0) \times (b_0: b_1: b_2: 0) \\[-8pt]
 \rotatebox{270}{$\mapsto$} \\ 
 (\underbrace{a_0b_0}_{z_{00}}: \underbrace{a_1b_1}_{z_{11}}: \underbrace{a_2b_2}_{z_{22}}: \underbrace{0}_{z_{33}}: \underbrace{a_0b_1 + a_1b_0}_{s_{01}}: \underbrace{a_0b_2 + a_2b_0}_{s_{02}}: \underbrace{0}_{s_{03}}: \underbrace{a_1b_2 + a_2b_1}_{s_{12}}: \underbrace{0}_{s_{13}}:\underbrace{0}_{s_{23}})
\end{gathered}
\]
We define $Y_{\pinv}$ to be the cone over this image (here ``$\pinv$'' stands for ``planar-involutions'').
So $Y_{\pinv}$ is contained in the~$\p^{6}$ with coordinates~$z_{00}$, $z_{11}$, $z_{22}$, $s_{01}$, $s_{02}$, $s_{12}$, and~$l$,
has dimension five, and is defined by the cubic equation
\[
 s_{01} s_{02} s_{12} - s_{12}^2 (z_{00} + z_{11} + z_{22}) + 4 z_{00} z_{11} z_{22} = 0 \,.
\]
The restriction to this~$\p^6$ of the symmetric bilinear sphere condition from \Cref{eq:symmetric_bilinear_sphere_condition} is:
\begin{equation}
\label{eq:planar_symmetric_bilinear_sphere_equation}
 lh + z_{00} r - 2 ( s_{01} x_1 +s_{02} x_2) - 2 (z_{11} m_{11} + z_{22} m_{22}) - 2 s_{12} m_{12} = 0 \,.
\end{equation}
This form determines a duality between the~$\p^6$ containing~$Y_{\pinv}$ and the~$\p^6$ with coordinates $m_{11}, m_{12}, m_{22}, x_1, x_2, r, h$.
We denote the projection of~$X_{\inv}$ to these coordinates by~$X_{\pinv}$.
This is a variety of dimension four and degree six in~$\p^6$.

Now the construction works as follows:
we take a general~$\p^2$ in the~$\p^{6}$ containing~$Y_{\pinv}$ and we intersect it with~$Y_{\inv}$.
This determines a plane cubic curve in~$Y_{\pinv}$.
The dual to this plane is a~$\p^3$, which intersects~$X_{\pinv}$ in a curve of degree six.
Therefore, in this way we obtain a line-symmetric mobile infinity-pod.

\begin{remark}
 The plane cubic~$L$ in~$Y_{\pinv}$ determined by this construction
 is covered via a $2 \colon 1$ map by a curve $\widehat{L} \subset \p^2 \times \p^2$
 that spans a~$\p^5$ and has degree six once embedded in~$\p^8$ via the Segre embedding.
 This sextic curve~$\widehat{L}$ has bidegree~$(3,3)$ in~$\p^2 \times \p^2$
 since $L$ is defined by three linear equations.
 Therefore, the projections of~$\widehat{L}$ on the planes where the base and the platform lie have both degree three,
 so we get that both the base and the platform lie on cubic curves.
 Let $\widehat{L}_{\text{base}}$ and $\widehat{L}_{\text{plat}}$ be these two curves;
 notice that these two curves are actually the same, but the two projection maps differ: one is a composition by the other
 and the involution that interchanges base and platform points.
 This matches with the observation by Bricard in \cite[Chapter~V]{Bricard1906}
 which says, as reported in \cite[Section~1.2.2]{Gallet2017}, that the involution that interchanges base and platform points,
 considered as an involution of the curve $D := \widehat{L}_{\text{base}}$, has the following property:
 the tangents to~$D$ at a base point~$a$ and its corresponding platform point~$b$ have a common intersection belonging to~$D$.
 This can be explained by the fact that~$D$ is a curve of genus one,
 therefore the involution mapping~$a$ to~$b$ is a translation by an element of order~two.
 Let $c \in D$ be such element. This means that, using the group law of~$D$,
 we have $b = a + c$. Now, let $t \in D$ be the point where the tangent to~$D$ at~$a$ intersects~$D$.
 In terms of the group law, we have $2a + t = 0$.
 Then, proving the observation by Bricard amounts to showing that $2b + t = 0$,
 which follows immediately once we recall that $2c = 0$.
 The projections $\widehat{L} \longrightarrow \widehat{L}_{\text{base}}$ and $\widehat{L} \longrightarrow \widehat{L}_{\text{plat}}$ 
 are isomorphisms of curves of genus one.
 The curve $L$ also has also genus one, but it is not isomorphic to these three curves;
 in fact, the $2 \colon 1$ cover $\widehat{L} \longrightarrow L$ must be unramified because of the Riemann-Hurwitz formula.
\end{remark}

\begin{remark}
 Our setting allows us to interpret the result by Borel in \cite[Case~Fb3]{Borel1908} which,
 as reported in \cite[Section~1.2.2]{Gallet2017}, reads as follows:
 in addition to the infinitely many legs having base and platform points lying on a planar cubic,
 there exist up to eight more legs which come in four symmetric pairs,
 where the symmetry is the same as for the other infinitely many legs.
 To recast this result in our framework,
 we use the connection between legs of line-symmetric pods and spectrahedra introduced in \cite[Section~4.2]{Gallet2017}.
 To stay faithful to the general spirit of the current paper,
 we do not introduce all the technical details,
 but rather we give a sketch that the interested reader can complement with the material in~\cite{Gallet2017}.
 We argue as follows: we have seen that
 the configuration curve~$C$ of a line-symmetric planar infinity-pod constructed in this section spans a~$\p^3$.
 This curve is contained in~$X_{\pinv}$ which, as we mentioned before, is a $2:1$ quotient of~$X_{\inv}$.
 The preimage~$\widetilde{C}$ of~$C$ under this $2:1$ map spans a~$\p^6$:
 in fact, the curve~$C$ is determined by three linear equations,
 which in the $\p^{10}$ where $X_{\inv}$ lives define a~$\p^7$;
 however, since $\widetilde{C}$ is contained in $X_{\inv}$ 
 and the latter satisfies a linear equation in~$\p^{10}$,
 we get that $\widetilde{C}$ spans a~$\p^6$.
 The dual of this~$\p^6$ is a~$\p^3$, which we denote by~$\Gamma$, in the~$\p^{10}$ containing~$Y_{\inv}$.
 The intersection of~$\Gamma$ with~$Y_{\inv}$ contains the planar cubic curve of legs that we described in this section.
 However, it contains something more.
 
 Once we identify the~$\p^{10}$ where $Y_{\inv}$ lives with the space of symmetric $4\times4$ matrices,
 the points in~$Y_{\inv}$ correspond to symmetric matrices of rank~two.
 Inside the linear space~$\Gamma$ there is a quartic surface, called \emph{symmetroid} and denoted~$\mscr{S}$,
 whose points correspond to symmetric matrices of rank at most~three.
 The points of~$Y_{\inv} \cap \Gamma$ are the singular points of~$\mscr{S}$.
 Since we know that $Y_{\inv} \cap \Gamma$ contains a planar cubic,
 it follows that $\mscr{S}$ must be reducible,
 more precisely the union of a plane and a cubic surface.
 The equation of~$\mscr{S}$ in the variables $w_0$, $w_1$, $w_2$, and~$w_3$ is given by
 \[
  \det( w_0 \, E + w_1 \, A_1 + w_2 \, A_2 + w_3 \, A_3 ) = 0 \,,
 \]
 where $E$, $A_1$, $A_2$, and $A_3$ span~$\Gamma$.
 From~\cite{Gallet2017}, we know that $E$ is
 \[
  \begin{pmatrix}
   0 & 0 & 0 & 0 \\
   0 & 1 & 0 & 0 \\
   0 & 0 & 1 & 0 \\
   0 & 0 & 0 & 1
  \end{pmatrix}
 \]
 and the matrices $A_i$ can be taken with zeros in the last row and column.
 The shapes of the matrices~$E$, $A_1$, $A_2$, and~$A_3$ force the determinant to look as follows:
 \[
  \det
  \begin{pmatrix}
   \ast & \ast & \ast & 0 \\
   \ast & \ast & \ast & 0 \\
   \ast & \ast & \ast & 0 \\
   0 & 0 & 0 & w_0
  \end{pmatrix}
  = w_0 H \,,
 \]
 where $H$ is a cubic polynomial in~$w_0, w_1, w_2, w_3$.
 Therefore, the symmetroid~$\mscr{S}$ is the union of the plane $w_0 = 0$
 and a cubic symmetroid~$\mscr{T}$ defined by $H = 0$.
 
 We recover the plane cubic curve of pairs of legs as the intersection $\{ w_0 = 0 \} \cap \mscr{T}$.
 However, these are not the only singular points of~$\mscr{S}$:
 we need to consider the singularities of~$\mscr{T}$.
 It is well known that cubic symmetroids has, in general, four nodes \cite{Ottem2015}, which may be all real.
 From this we see that we may have up to additional four pairs of legs for the infinity-pod,
 which matches the result by Borel.
\end{remark}

\section{A new construction of infinity-pods}
\label{construction}

After getting acquainted with the projective models of isometries and legs in the previous sections,
we now come to the main result of this paper, namely, a new construction (to our knowledge) of line-symmetric mobile infinity-pods.

The starting point for the construction is the remark that, if a pod has configuration curve~$C$
contained in~$X_{\inv}$ whose linear span has dimension five then the legs compatible under the symmetric bilinear sphere condition
are given by the intersection of~$Y_{\inv}$ with a linear space of dimension four.
Since $Y_{\inv} \subset \p^{10}$ has dimension seven, the latter intersection is at least a curve, 
so the pod is mobile and admits infinitely many legs.
In addition, we will ensure that the infinity-pod we construct has a single degree of freedom and not more.

Our goal, then, becomes to construct a curve contained in~$X_{\inv}$ whose linear span has dimension five.
To do so, we start with the rotational part of the desired configuration curve, namely, 
with a curve in the~$\p^3$ with so-called Euler coordinates $(e_0: e_1: e_2: e_3)$.
We then lift this curve to~$X_{\inv}$ in~$\p^{16}$.
From \Cref{projective_models}, we know that such a curve must satisfy $e_0 = 0$.
Therefore we start from a plane curve
\[
 C \subset \{e_0 = 0\} \cong \p^2_{(e_1:e_2:e_3)}
\]
of equation $F(e_1, e_2, e_3) = 0$.
We exclude curves~$C$ of degree at most two because their behavior is well known; see~\cite{Husty2002}. 
Then, the image of~$C$ under the Veronese map from \Cref{eq:veronese} spans a~$\p^5$.
Therefore, if we want the lift~$\widetilde{C}$ of~$C$ to~$\p^{16}$ to span a~$\p^5$
then the restriction to the span of~$\widetilde{C}$ of the projection $X \dashrightarrow X_{\mathrm{rot}}$ must be a linear isomorphism.
In other words, the $x$-, $y$-, and $r$-coordinates of the points of~$\widetilde{C}$ must each be linearly dependent from the $(M, h)$-coordinates.

We first lift the curve~$C$ to $Z_{\inv} \subset \p(\vec{1}, \vec{2})$ from \Cref{eq:Zinv}
and then we lift it further to~$\p^{16}$.
Lifting the curve~$C$ to~$Z_{\inv}$ amounts to finding three polynomials
\[
 P_1, P_2, P_3 \in \R[e_1, e_2, e_3] \,,
\]
which yield the map
\[
\begin{array}{rcl}
 C & \longrightarrow & Z_{\inv} \\
 (0: e_1: e_2: e_3) & \mapsto & (e_1: e_2: e_3: \underbrace{P_1: P_2: P_3}_{p}: \underbrace{0: 0: 0}_{q})
\end{array}
\]
Because of the equations defining~$Z_{\inv}$, we know that the polynomials $P_1, P_2, P_3$ must satisfy
\[
 e_1 P_1 + e_2 P_2 + e_3 P_3 = 0
\]
modulo the equation~$F$ of the curve~$C$.
In other words, the triple $(P_1, P_2, P_3)$ must be a syzygy of~$(e_1, e_2, e_3)$ modulo~$F$.
We hence must have
\begin{equation}
\label{equation:syzygy}
 \begin{pmatrix}
  P_1 \\ P_2 \\ P_3
 \end{pmatrix}
 = 
 L_1
 \begin{pmatrix}
  0 \\ -e_3 \\ e_2 
 \end{pmatrix}
 + 
 L_2
 \begin{pmatrix}
  e_3 \\ 0 \\ -e_1
 \end{pmatrix}
 +
 L_3
 \begin{pmatrix}
  -e_2 \\ e_1 \\ 0
 \end{pmatrix} \mod F \,.
\end{equation}
Because of the connection between the $p$-variables and the $x$- and $y$-variables,
we know that the $p$-variables should be linearly dependent from the span of quadratic monomials in $e_1, e_2, e_3$.
Therefore, $P_1$, $P_2$, and~$P_3$ should be quadratic polynomials, namely in the previous equation
we take~$L_i$ to be linear in~$e_1$, $e_2$, $e_3$.
In turn, this means that either $F = e_1 P_1 + e_2 P_2 + e_3 P_3$, or $e_1 P_1 + e_2 P_2 + e_3 P_3$ is the zero polynomial.
However, considerations from \cite{Gallet2015} imply that if $F$ were of degree three, 
then colinearities would have to happen among base or platform points.
We do not consider these cases particularly interesting, therefore, 
from now on we suppose $F$ to have degree higher than three.
This means that $e_1 P_1 + e_2 P_2 + e_3 P_3 = 0$. 

With these choices, we see that the points in the lift~$\widetilde{C}$ must satisfy 
\[
 (e_1^2 + e_2^2 + e_3^2) r = \frac{1}{4} (P_1^2 + P_2^2 + P_3^2) \,,
\]
because $h = e_1^2 + e_2^2 + e_3^2$ and $x_i = P_i/2$.
However, let us look at the condition imposed by a leg with base point~$a$, platform point~$b$, and length~$d$ on the points in~$X$:
\[
 \left\langle x, a \right\rangle + \left\langle y, b \right\rangle + \left\langle Ma, b \right\rangle - \frac{1}{2} r - \left\langle a, b \right\rangle h = 0 \,.
\]
We see that the variable $r$ is hence linearly dependent from the other variables for the points of a configuration curve of a pod.
Therefore, we must have that (see also \cite[Theorem~7.9]{GrafVonBothmer2020})
\[
 (e_1^2 + e_2^2 + e_3^2) \text{ divides } (P_1^2 + P_2^2 + P_3^2) \mod F \,.
\]
If $P_1^2 + P_2^2 + P_3^2$ were a multiple of~$e_1^2 + e_2^2 + e_3^2$ already in~$\C[e_1, e_2, e_3]$,
we could lift the whole plane $\{ e_0 = 0 \} \subset \p^3$ to~$\p^{16}$.
Therefore, we would obtain a pod whose configuration space is two-dimensional, and not one-dimensional.
We do not want this situation, so we must have
\[
 P_1^2 + P_2^2 + P_3^2 = U (e_1^2 + e_2^2 + e_3^2) + VF \,,
\]
where $U$ is a quadratic polynomial, and $V$ is a polynomial of degree at most~$1$.
Since we do not want $F$ to be of degree three,
the curve~$C$ must be a plane quartic.

On the leg side, the dual of span of the curve~$C$ is a $\p^4$.
Intersecting this~$\p^4$ with~$Y_{\inv}$ give a curve~$L$ parametrizing infinitely many pairs of legs.
We can compute the Hilbert series of~$L$ in terms of the Hilbert series of~$Y_{\inv}$;
we get that the degree of~$L$ is~ten (the same as the degree of~$Y_{\inv}$) and its genus is~six.

Our construction is explained in Algorithm~\texttt{CreateInfinityPod}.

\renewcommand{\thealgorithm}{} 
\begin{algorithm}[ht]
\caption{\texttt{CreateInfinityPod}}\label{alg:CountRealizations}
{\small
\begin{algorithmic}[1]
  \Require None.
  \Ensure The configuration curve and the leg variety of a line-symmetric infinity pod.
  \Statex \vspace{-8pt}
  \State {\bfseries Pick} three random linear forms $L_1, L_2, L_3 \in \C[e_1, e_2, e_3]$.
  \State {\bfseries Pick} a random quadratic form $U \in \C[e_1, e_2, e_3]$.
  \State {\bfseries Define} three quadratic monomials $P_1, P_2, P_3 \in \C[e_1, e_2, e_3]$ by
   \[
    \begin{pmatrix}
     P_1 \\ P_2 \\ P_3
    \end{pmatrix}
    := 
    L_1
    \begin{pmatrix}
     0 \\ -e_3 \\ e_2 
    \end{pmatrix}
    + 
    L_2
    \begin{pmatrix}
     e_3 \\ 0 \\ -e_1 
    \end{pmatrix}
    +
    L_3
    \begin{pmatrix}
     -e_2 \\ e_1 \\ 0
    \end{pmatrix}  \,.
   \]
  \State {\bfseries Define} $F := (P_1^2 + P_2^2 + P_3^2) - U(e_1^2 + e_2^2 + e_3^2)$.
  \State {\bfseries Define} the ring homomorphism
   \[
    \begin{array}{rrcl}
     \rho \colon& \C[M, x, y, r, h] & \longrightarrow & \C[e_1, e_2, e_3] \\
     & (m_{11}, m_{12}, \dotsc, m_{33}) & \mapsto & 
     \arraycolsep=1pt
     \begin{array}{rll} 
      ( &e_{1}^{2} -  e_{2}^{2} -  e_{3}^{2}, 2 e_{1} e_{2}, 2 e_{1} e_{3}, \\
        &2 e_{1} e_{2}, -  e_{1}^{2} + e_{2}^{2} -  e_{3}^{2}, 2 e_{2} e_{3}, \\
        &2 e_{1} e_{3}, 2 e_{2} e_{3}, -  e_{1}^{2} -  e_{2}^{2} + e_{3}^{2}&)
     \end{array} \\
     & (x_1, x_2, x_3) & \mapsto & \frac{1}{2} (P_1, P_2, P_3) \\
     & (y_1, y_2, y_3) & \mapsto & \frac{1}{2} (P_1, P_2, P_3) \\
     & r & \mapsto & U \\
     & h & \mapsto & e_1^2 + e_2^2 + e_3^2 
    \end{array} 
   \]
  \State {\bfseries Compute} the ideal~$J_{\inv}$ of~$X_{\inv}$.
  \State {\bfseries Define} $I := J_{\inv} +$  (preimage of the ideal~$(F)$ under~$\rho$).
  \State {\bfseries Compute} the vector space~$I_{\mathrm{lin}}$ of linear polynomials in~$I$.
  \State {\bfseries Compute} the vector space~$L_{\mathrm{lin}}$ dual to~$I_{\mathrm{lin}}$ with respect to the bilinear form
   \[
      l  h + z_{00} r  - 2  ( z_{10} x_1 + z_{20} x_2 + z_{30} x_3 ) 
     - 2  (z_{01} y_1 + z_{02} y_2 + z_{03} y_3 ) - 2  \sum_{i, j = 1}^{3} 
    m_{ij}  z_{ij} =  0 \,.
   \]
  \State {\bfseries Compute} the ideal~$N$ of~$Y$.
  \State {\bfseries Define} $L := N + L_{\mathrm{lin}}$.
  \State \Return $(I, L)$.
\end{algorithmic}}
\end{algorithm}

\begin{remark}
The curve~$L$ is a $2:1$ image of a curve $\widetilde{L} \subset Y$ parametrizing infinitely many legs. 
This is a curve of degree~$20$ spanning a $\p^{10}$, which can be obtained by intersecting~$Y$ with six linear forms. 
The Hilbert series shows that this curve has genus~$11$, hence it is a canonical curve. 
\[
 \xymatrix{
  \p^{10} \ar@{}[r]|-{\supset} \ar@{}[d]|-{\rotatebox{90}{$\subset$}} & Y_{\inv} \ar@{}[d]|-{\rotatebox{90}{$\subset$}} \\
  \p^4 \ar@{}[r]|-{\supset} & L & \widetilde{L} \ar[l]_{2:1} \ar@{}[r]|-{\subset} & \p^{10}
 }
\]
By projecting each leg to its base point, we can isomorphically project $\widetilde{L}$ to a curve $\widetilde{L}_{\text{base}} \subset \p^3$, which has degree~ten.
The projection to the platform point gives the same curve, but the two projections differ by an involution of~$\widetilde{L}_{\text{base}}$.
\end{remark}

\end{document}